# In-Situ Infrared Camera Monitoring for Defect and Anomaly Detection in Laser Powder Bed Fusion: Calibration, Data Mapping, and Feature Extraction

Shawn Hinnebusch, David Anderson, Berkay Bostan, and Albert C. To*

Department of Mechanical Engineering and Materials Science, University of Pittsburgh, Benedum Hall, 3700 O'Hara Street, Pittsburgh, PA 15261, USA

* Corresponding author: albertto@pitt.edu

**Abstract**

Laser powder bed fusion (LPBF) process can incur defects due to melt pool instabilities, spattering, temperature increase, and powder spread anomalies. Identifying defects through in-situ monitoring typically requires collecting, storing, and analyzing large amounts of data generated. The first goal of this work is to propose a new approach to accurately map in-situ data to a three-dimensional (3D) geometry, aiming to reduce the amount of storage. The second goal of this work is to introduce several new IR features for defect detection or process model calibration, which include laser scan order, local preheat temperature, maximum pre-laser scanning temperature, and number of spatters generated locally and their landing locations. For completeness, processing of other common IR features, such as interpass temperature, heat intensity, cooling rates, and melt pool area, are also presented with the underlying algorithm and Python implementation. A number of different parts are printed, monitored, and characterized to provide evidence of process defects and anomalies that different IR features are capable of detecting.

*Keywords*: Additive manufacturing; Laser powder bed fusion; In-situ monitoring; Defect detection

# 1 Introduction

Over the past decade, additive manufacturing (AM) has experienced significant growth, particularly as technology advances toward large-scale production [1], [2]. Among the various AM techniques, laser powder bed fusion (LPBF) is known for its ability to produce high-density parts, surpassing methods like binder jetting. Despite achieving over 99.5% density in parts fabricated using commercially available materials, LPBF still generates defects stemming from the manufacturing process due to temperature accumulation, spatter generation, powder spread anomalies, and melt pool instabilities [3], [4]. These defects can lead to premature part failure and pose significant risks, especially in critical components such as those used in aerospace applications.

There is a growing emphasis on in-situ monitoring as a critical step in mitigating build failures and qualifying parts. For example, Baumgartl et al. [5] employed a machine learning algorithm to detect spatter and delamination from thermographic off axis images. Schwerz et al. [3] used the EOS 5-megapixel sCMOS (scientific complementary metal-oxide-semiconductor) camera using the Laplacian of Gaussian then identified local minima for spatter. McNeil et al. [6] used in-situ infrared (IR) and optical imaging to provide a pathway to defection detection, mainly using IR thermal signatures of intensity and thermal decay. Lough et al. [7] used a short-wave infrared (SWIR) camera and extracted features, including the apparent melt pool area, time above threshold, maximum radiance, maximum radiance decrease rate, and radiance sum above threshold, then combined the data with CT into a voxel reconstruction. Lang et al. [8] published a paper generating 36 synthetic datasets from the original 3 in-situ features. Wang et al. [9] used an open-source machine to collect co-axial melt pool data to regulate the melt pool geometry using spatial iterative learning control (SILC). Oster et al. [10] used an off-axis SWIR camera with a frame rate of 2192 Hz, FOV approximately 10x10 mm, and 38 µm spatial resolution in their work, where they generated 23 different melt pool related features and combined them with XCT data for machine learning defect detections. Guirguis et al. [11] reconstructed the CAD geometry during post-processing.

Reducing large amounts of in-situ data into several key features is a crucial step toward minimizing memory requirements for machine learning and overall storage needs. Thus the first goal of this work aims to introduce new IR features for defect detection or process model calibration together with evidence of defects for a corresponding feature if applicable. The new IR features that will be introduced include laser scan order, local preheat temperature, number of spatters generated locally, and spatter landing locations. In particular, tracking landing locations of the spatters is crucial because they are often responsible for lack-of-fusion defects [12]. Lack-of-fusion defects are large irregular defects that can be formed from incomplete meting, often related to large spatter particles. Although melt pool size and number of spatters generated locally have been presented as features for in-situ monitoring based on high-speed cameras [10], the processing algorithms for obtaining them from IR images have not been published yet. For completeness, processing for other common IR features, such as interpass temperature, heat intensity, and cooling rates, will also be discussed in this work.

The second goal of this work is to introduce a new approach to accurately map in-situ data to a three-dimensional (3D) geometry model, represented in the stereolithography (STL) format, aiming to reduce storage requirements and pave the way for real-time defect detection. Existing methods for in-situ mapping are typically a manual process involving thresholding the results and generating a point cloud for alignment with the 3D geometry [13], [14]. While effective for post-processing results, this approach has limitations, particularly with complex geometries where different features or parts may exhibit significant temperature variations. Moreover, real-time detection is constrained by the need for post-build alignment to the 3D model. Although 3D voxelization methods have been proposed [15], there is a lack of detailed information regarding the mapping process to the 3D model. This work proposes an approach that corrects the image distortion and places each pixel into the correct location on the 3D geometry without the need for any post-processing. The image registration in this approach provides details on the specific pixels located inside the part before beginning the build removing the tedious task of alignment. This approach automates the selection and placement of each pixel within the part while significantly reducing storage space requirements, which would enable future real-time defect detection.

The paper is structured as follows: Section 2 details the experimental setup and thermal calibration of the infrared (IR) camera for several temperatures and surface conditions. In Section 3, we outline the custom plate design for angle and distortion correction, along with the novel mapping technique employed to project thermal images onto complex geometries. Section 4 presents 10 distinct IR features developed to highlight potential defects and provide valuable temperature profiles beneficial for simulations. Finally, Section 5 concludes with a summary of the study and presents the key takeaways.

## 2 Infrared Camera Mesh Projection Mapping

### 2.1 Experimental Setup

The L-PBF system (EOS M290 DMLS) used in this work is equipped with a FLIR A700 infrared camera with a 640x480 pixel detector and a 24° lens capable of recording up to 30 frames per second (fps) shown in Figure 1. The camera remains a fixed distance to the part as the substrate lowers and new powder is spread. FLIR Research Studio is utilized for all the recordings and for viewing the real-time progress of the build. A simple switch is installed near the recoat blade to trigger the stop/start of the recordings for each layer. The camera offers three temperature ranges: -20 °C to 120 °C, 0 °C to 650 °C, and 300 °C to 2000

°C. A custom-built heating module designed for the EOS machine provides a preheat temperature of 500 °C on a 4×4 inch plate. Utilizing this custom heating module in conjunction with thermocouples, the camera is calibrated for various scanning strategies and surface conditions, ensuring accurate measurements across temperatures ranging from room temperature to 500 °C. Current camera capabilities include angle perspective corrections to compensate for camera angle distortions and any rotations resulting from camera placement. Following image correction, a voxel mesh of the geometry is superimposed on the infrared data to distinguish between the powdered surface and the part. A 3D reconstruction of the part identifies locations with high heat accumulation, laser intensity, spatter generation and landing location, surface roughness, powder spread, and cooling rates to pinpoint local areas of defect anomalies. Inconel 718 (IN 718) printed with the default EOS process parameters (Table 1) is used to demonstrate the IR data processing algorithms throughout this work.

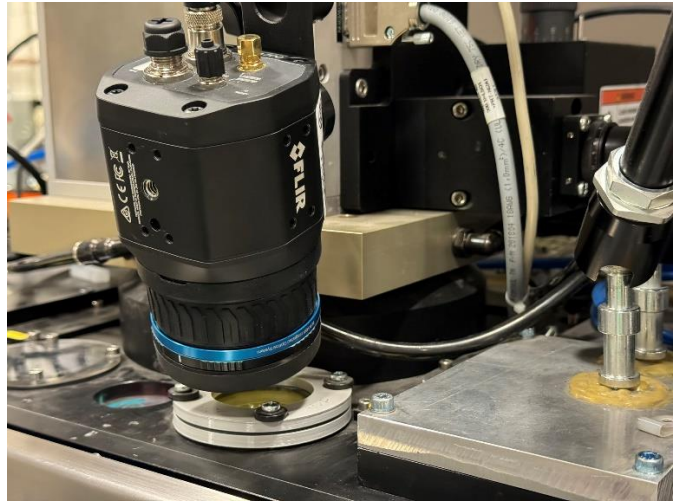

Figure 1: Experimental setup of the FLIR A700 camera mounted on the EOS M290 DMLS system

Table 1: Default LPBF process parameters for Inconel 718 in EOS M290 DMLS system

| Process Parameter | Value |
| --- | --- |
| Layer thickness | 40 µm |
| Hatch spacing | 110 µm |
| Stripe width | 10 mm |
| Stripe overlap | 0.08 mm |
| Laser rotation angle | 66.7° |
| Infill | Laser Power 285 W<br>Scan Speed 960 mm/s |
| Contour 1 | Laser Power 138 W<br>Scan Speed 300 mm/s |
| Contour 2 | Laser Power 80 W<br>Scan Speed 800 mm/s |
| UpSkin | Laser Power 153 W<br>Scan Speed 600 mm/s |
| DownSkin | Laser Power 145 W<br>Scan Speed 2400 mm/s |
| Platform Temperature | 80 °C |

## 2.2 Temperature Calibration

Accurately measuring temperature using an IR camera presents challenges due to variations in emissivity caused by temperature fluctuations, material composition, and surface roughness. Without proper calibration of the emissivity value, estimating true temperature values is not feasible [16]. Block body calibration methods are commonly employed to establish the correlation between camera signals and temperature [17], [18]. This study performs a comprehensive calibration procedure using different scanning techniques on IN718 material.

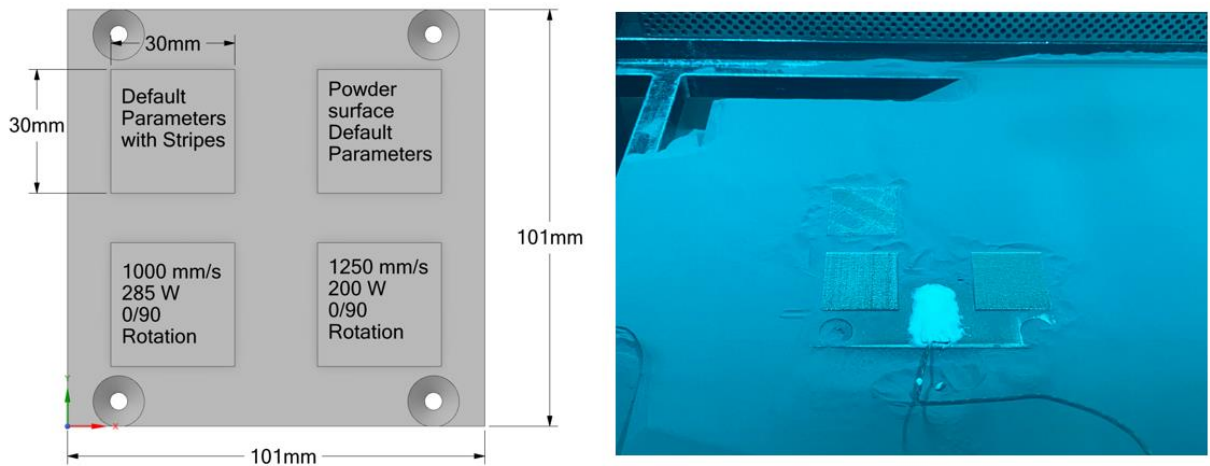

Figure 2: Temperature calibration build for IR emissivity using as-printed surfaces for default process parameters with stripes (upper-left block), EOS M290 default process parameters with 0/90 pattern rotation and no stripes (lower-left block), process parameters in the conduction regime with 0/90 pattern rotation but no stripes (lower-right block), and powder surface (upper-right block).

For this experiment, four 30×30-mm blocks are printed on a square 4-inch by 3/16-inch thick plate mounted on the custom heating module. Each block is printed to a height of 1 mm (comprising 25 layers) to ensure the powder layer thickness reaches steady state before finishing the build. During the final recoat phase, one block (top right block on the plate shown in Figure 2) is intentionally left with powder while the remaining three blocks are scanned. The powder is carefully removed to mount a Type K thermocouple with adhesive. Temperatures are acquired with a National Instruments data acquisition device (DAQ) ranging from room temperature up to 500 °C.

During the heating module operation, the top surface exhibits a reduced temperature while reaching the maximum temperature of 500 °C. As the thermal camera is mounted outside the machine, a 50-mm diameter by 3-mm thick Zinc Selenide (ZnSe) glass viewport window is installed, and an additional correction factor is calibrated to account for the signal alteration. Data collection is performed with and without the ZnSe glass to determine its transmission rate through the lens. Temperature increments of 100 °C from room temperature up to 300 °C are used to calculate the same temperature with and without the glass by changing the transmission factor. A glass transmission factor of 0.75 provides consistent temperature readings compared to those obtained without the glass. The machine is purged with Argon gas for calibration completed above 300 °C with the ZnSe glass remaining on for the remaining experiments.

The IR surface emissivity is adjusted to reduce the temperature error from the camera and the thermocouple. Figure 3(a) illustrates an image of the raw data during the calibration. Although the temperatures of the four blocks are close in temperature, the temperatures on the as-printed surfaces are less than that of the powder. This is due to the emissivity difference and must be corrected to obtain accurate temperature measurements. Figure 3(b) compares the temperatures and standard deviation, after the emissivity correction, of the 3 as-printed surface conditions and the powder surface with the thermocouple temperatures. The calibrated emissivity value remains constant across the temperature range up to 500 °C, consistent with findings reported in existing literature [19]. The powder surface exhibits the most consistent temperature readings, average of 3.3 °C standard deviation and less than 5 °C maximum average error, with an emissivity value of 0.63. As-printed surfaces result in a higher deviation of 25.2 °C throughout the surface and have a much lower emissivity value of 0.21 with a 20 °C maximum average error. As-printed surfaces with stripes typically have the highest temperature variation across all scanning strategies tested. Previous studies by Del Campo et al. [20] demonstrated that for aeronautical alloys, including unoxidized Inconel 718, emissivity typically falls between 0.2 and 0.7, remaining nearly independent of temperature (200 to 650 °C), which aligns with our findings here.

Comparative analysis reveals that the 0/90 scanning strategy without stripes yielded lower temperature deviation compared to surfaces with stripes. Notably, determining temperature on as-printed striped surfaces results in the highest error due to surface roughness. Extracted features rely on the powder surfaces for the most accurate readings presented in this work, with additional features leveraging the high deviation to aid in defect detection.

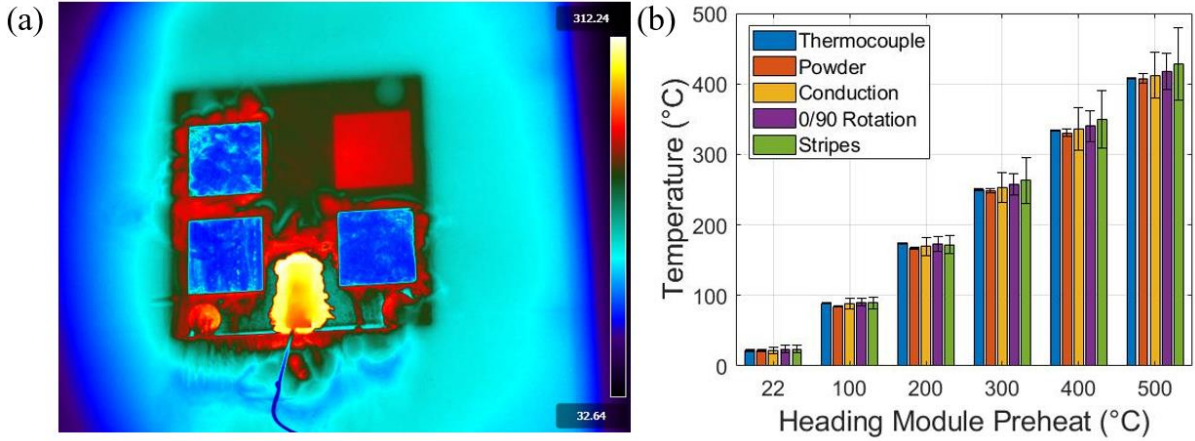

Figure 3: Temperature calibration of IR data for different surfaces: (a) Raw IR image displaying temperature outcomes for the four surface printing conditions, and (b) average temperature and standard deviation of each block at various heating module temperatures. The calibrated surfaces depicted in Figure 2 are then compared to the thermocouple measurements. Notatatons: Powder (upper-right block), Conduction (lower-right block), 0/90 Rotation (lower-left block), and Stripes (upper-left block).

## 3 Mesh Projection Mapping

### 3.1 Spatial Calibration

The laser and optics, positioned directly above the build platform, necessitate mounting the IR camera in a viewport at an angle. The camera is positioned on a movable arm which alters location and camera angle with each mounting, posing challenges in aligning parts accurately according to the 3D geometry. To address this issue, a custom calibration plate is fabricated for the machine to facilitate alignment and resolve the camera position upon mounting. Built from a ¼" think MC6 cast aluminum sheet, the plate contains 1-mm through-holes drilled to easily highlight pixel locations. The holes are positioned at the four corners and midpoints of the three squares (50, 100, and 150 mm), enabling easy pinpointing of the exact locations in the printing area. Additionally, a center hole aids in aligning the plate to the 3D geometry for all the geometry offsets. Four legs were machined to fit into each of the build plate's build holes to ensure proper placement. As the IR camera only captures thermal temperatures, the plate is heated in a furnace to 50 °C for easy visibility of the holes and then placed in the EOS machine. The plate is lowered to the correct printing height and checked for recoater blade interference before taking a calibration image with the results depicted in Figure 4. Raw image discrepancies intensify near the 150-mm square due to the camera's angle and rotation, resulting in over a 10-mm error between the expected and actual locations of the plate.

To mitigate this error, an image is captured before the build and is utilized for perspective correction, considering the camera's angle and rotation. This correction is implemented via a Python-based code developed using the OpenCV [21] command *getPerspectiveTransformation()* based on homography transformation [25]. Selecting the (*x,y*) coordinates of four points on the 150-mm square plane, a 3×3 homography matrix **H** is calculated based on Equation (1) for mapping any points on the image to the correct location as follows:

$$\begin{bmatrix} x' \\ y' \\ 1 \end{bmatrix} = \mathbf{H} \begin{bmatrix} x \\ y \\ 1 \end{bmatrix} \qquad (1)$$

where $x'$ and $y'$ are the corrected image coordinates. Hence once the components in the matrix **H** have been determined, the equations can be used in *getPerspectiveTransform()* to correct all images throughout the build.

With the proposed image correction method, pixel locations are determined from the center of the build plate and easily mapped to the 3D model. Pixel resolution is calculated at 360 microns by counting the number of pixels in a known length near the center of the image. The previously high error of 10 mm is minimized to the nearest pixel or approximately ±360 µm of uncertainty.

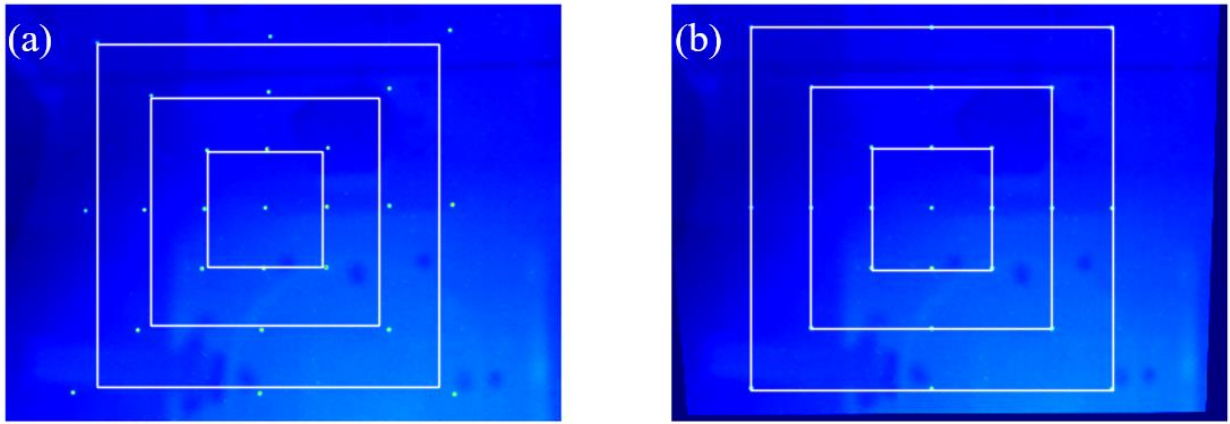

Figure 4: Demonstration of perspective correction using known locations on the build substrate: (a) Raw distorted image and (b) corrected image. The image was corrected using the 150 mm (outermost) square, and the positions were validated with the 100 and 50 mm squares. The middle dot helps locate the center of the substrate.

## 3.2 Voxel Mesh Generation and Projection

The process of mapping in-situ features back into the 3D geometry poses significant challenges, especially for complex geometries. McNeil et al. [6] mentioned the need for registration data from in-situ monitoring back into the 3D model as there are no commercial tools available and tried to use an adaptive thresholding. Although adaptive thresholds can provide reasonably accurate results for some geometries, these algorithms struggle for more complex geometries, as the local intensity or temperature measured can vary significantly depending on the geometry and location within the part. This creates a need to develop different features and reconstruct them back onto the true 3D geometry.

Registering pixels to the 3D geometry is a crucial step in achieving precise placement accuracy. The proposed mapping approach (Figure 5) begins by discretizing the geometry represented in the STL format into a voxel mesh that aligns with the resolution of the camera. Each voxel in the mesh is assigned a layer attribute based on the layer thickness to streamline post-build data processing. Every voxel in the mesh corresponds directly to a pixel in the image-corrected results to ensure an exact match with the 3D geometry. By retaining only data within the geometry, the tens to hundreds gigabytes of raw data are drastically reduced to just a couple of hundred megabytes in size. This represents a remarkable reduction in file size, often exceeding 99% in some instances.

The PyVista Python package is employed to perform mesh voxelization of the 3D geometry while aligning it to the 360-micron resolution of the IR camera. The PyVista package is an open-source 3D visualization tool with a Python interface, which is a high-level API to the Visualization Toolkit (VTK) [22]. The meshing software is based on *pyvista.voxelize* function where the mesh density of [x,y,z] provides a mesh of the same mesh resolution in the x/y direction and the build layer thickness. Accordingly, the STL file is meshed using a voxel size of $360 \times 360 \times 40 \ \mu m^3$. After generating the mesh for an STL or multiple STL files, the mesh is then used for in-situ processing. By generating the voxel mesh ahead of the build, real-time detection is possible as each pixel needed layer by layer is already known without the need for adaptive thresholding.

Once the voxel mesh is generated, each IR feature is then mapped onto every voxel layer in succession. Leveraging the knowledge of the build platform's center location (Figure 4), and the geometry's center location, the two origins (0,0) can be located. Given a camera resolution of 360 µm, the following positions (0, 360) and (0, 720) are mapped to the respective pixel coordinates of (0, 1) and (0, 2). Feature pixels are recorded onto the voxel mesh if the geometry is present. This ensures that the quantities of interest in the whole part are captured for subsequent processing and statistical analysis especially for complex geometries. The lower-right panel in Figure 5 presents a significant temperature disparity between the bottom (furthest from gas flow) and the top of the part, a contrast undetected through small-scale sampling.

The reconstructed part facilitates the establishment of thresholds for interpass temperature and heat map, enabling precise identification of locations prone to defect formation.

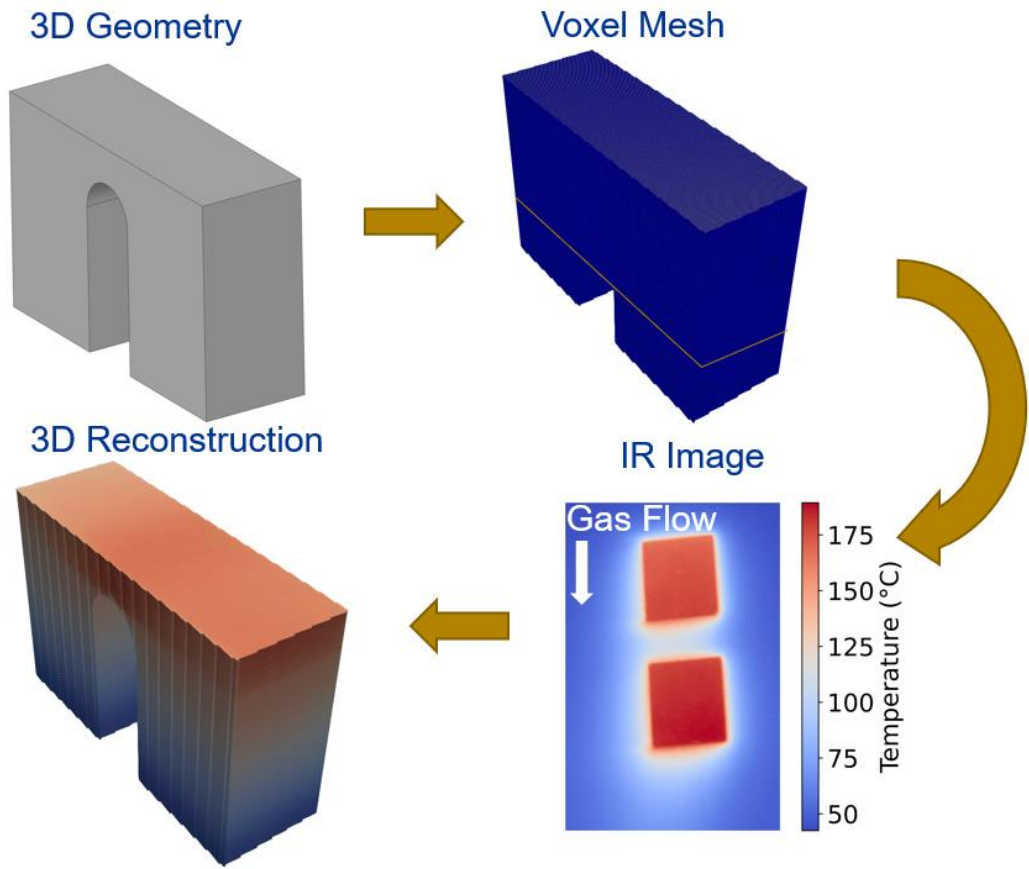

Figure 5: Proposed IR camera mesh projection workflow for 3D geometry of an arch structure: (step 1) Volexize the geometry, (step 2) process the raw IR data to obtain certain feature (e.g. interpass temperature), and (step 3) map the feature onto the voxel mesh.

To test the robustness of the algorithm, the Qualification Test Artifact (QTA) shown Figure 6(a) is selected for its geometric complexity. The QTA block was developed by Taylor et al. [23] which comprise intricate features such as overhangs, lattice structures, and embedded tensile bars – a comprehensive qualification test in a single part. Figure 6(b) is a 2D slice of the geometry on layer 463 in the block that is monitored by the IR camera. Figure 6(c) shows the corresponding interpass temperature mapped to the part without the powder bed using the proposed algorithm. Note that all the fine details of the solid regions in the part are visible. This example demonstrates the robustness of the mapping even for very complex geometries.

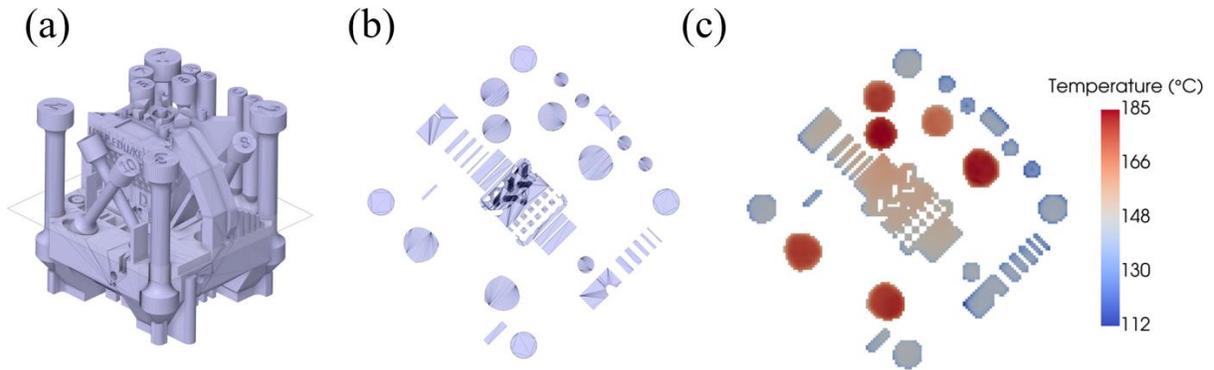

Figure 6: Demonstration of the proposed approach for mapping an IR feature onto a layer of the part: (a) The Qualification Test Artifact [23] with a 2D cutting plane at Layer 463, (b) STL 2D slice at Layer 463 representing the true geometry at the printing layer, (c) Interpass temperature on the same layer obtained by mapping the temperature onto the correct locations using the proposed approach.

## 4 IR Features

Storing the complete frame and every single image throughout the build process is impractical due to the immense file size and the challenges associated with processing such massive amounts of data for defect detection. Hence there is a need for streamlining the workflow and consolidating data into a manageable format. Additionally, machine learning algorithms must process in-situ data into memory for training purposes and must be compatible with defect identification methods such as cross-sectional imaging or micro-computed tomography (micro-CT). To address these issues, useful features are extracted from the raw IR data for capturing information about defects. These features can then be integrated with ML and processed to correct defects in real time with future developments. All the IR features extracted with a brief description are listed in Table 2 and will be discussed in detail below.

Table 2: Overview of IR Features

| IR Feature | Description |
| --- | --- |
| Interpass Temperature | The top surface temperature before the laser begins scanning the next layer. This feature can identify heat accumulation locations within components. |
| Heat Intensity | The maximum intensity (with emissivity set to 1.0) of the laser aggregated across all frames within the layer. This feature can identify significant intensity spikes caused by overhangs or substantial spatter occurrences. |
| Laser Scan Order | Provides an estimate of the scan sequence within the layer, which is usually unknown in commercial systems. It provides laser scanning time information for many additional features. |
| Local Pre-deposition Temperature | This feature is the temperature 0.33 seconds before the laser scans locally and can be used to capture heat accumulation within a layer. |
| Maximum Pre-deposition Temperature | This feature extends the concept of local pre-deposition temperature by identifying the highest temperature before the laser scans locally. This feature can inform spatter landing locations prior to the laser scanning locally in the same layer. |
| Spatter Generation | Determines the number of spatters (spatter count) ejected from the melt pool during laser rastering of each pixel, offering insights into melt pool instabilities locally. |
| Spatter Landing Location | Complementary to the maximum pre-deposition temperature, this feature identifies the spatter landing location by tracking all the sharp temperature changes between image frames. This feature can be used to identify lack-of-fusion defects generated by large spatters. |
| Relative Melt Pool Area | This feature provides a local melt pool area defined by a certain threshold temperature. Large melt pool fluctuations can lead to defects with unstable melt pool sizes. |
| Cooling Rate | While temperatures within the melt pool cannot be captured, the IR camera captures relative cooling rates. |
| Laplacian of Temperature | The Laplacian is applied to the interpass temperature to emphasize significant temperature variations, often associated with recoating issues. |
| As-printed Laplacian | This feature utilizes the as-printed surface temperature and applies the Laplacian to identify potential surface roughness resulting from changes in emissivity. |

## 4.1 Interpass temperature

The interpass temperature, also known as the inter-layer temperature (ILT) or the end-of-cycle temperature [24], [25]. It is defined as the temperature after recoating before the laser scanning. It provides information on the heat accumulation as the build continues through the print layers. The interpass temperature mediates melt pool geometry, defect generation, and microstructure evolution. For example, Chen et al. [26] explored the effects of preheat temperatures 100–500 °C on melt pool morphology in IN718 and found a transition in melt pool behavior as temperature increases . Olleak et al. [27] employed part-scale scan-resolved process modeling to simulate the heat transfer of an inverted pyramid and found a close correlation between interpass temperature and various phases in Ti6Al4V. Templeton et al. [28] identified shrinkage porosity at high energy densities, which can occur near standard process parameters at elevated pre-deposition temperatures. Figure 7 illustrates how shrinkage porosity varies with laser power, scanning velocity, and interpass temperature. Thus, capturing interpass temperature in parts could provide insights into defect detection and prevention.

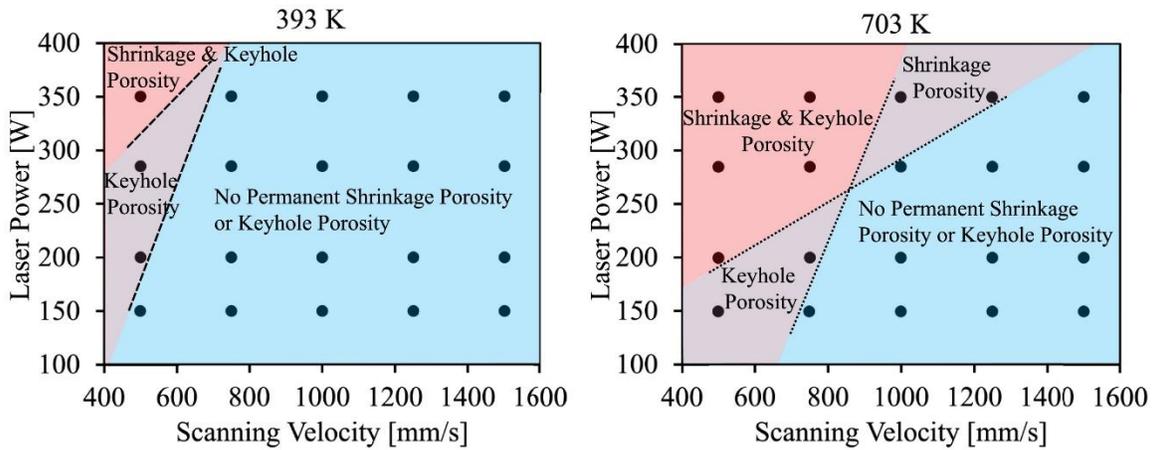

Figure 7: Process (power-velocity) map at (left panel) low interpass temperature and (right panel) high interpass temperature, illustrating a shift in the shrinkage and keyhole boundary conditions with increasing temperatures [28].

Figure 8 displays the interpass temperature distribution for layer 840 during the printing of the QTA block. Notably, this figure shows substantial temperature variations within a single layer during printing of a part. The colder columns or circular regions along the boundaries of the layer contrast sharply with the warmer central area, where the temperature difference can be as high as 188 ºC. The mapping scheme correctly identifies all features across the part.

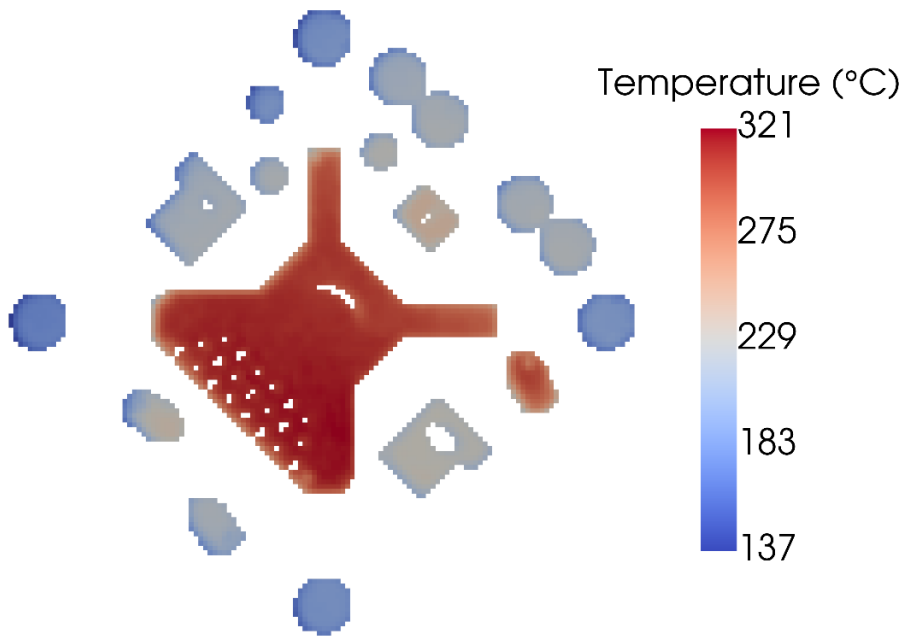

Figure 8: Interpass temperature profile on Layer 840 of the QTA Block

## 4.2 Heat Intensity Map

Although the resolution of our IR camera cannot capture detailed information within the melt pool, the relative intensity trend can be captured. The maximum temperature range of the IR camera in the mode we employed is 650 °C; however, because of the resolution and frame rate, the camera count (the maximum value a pixel can take on regardless of emissivity) does not saturate, thus allowing us to capture relative changes in the intensity. Figure 9 illustrates the heat intensity map (or heat map) on an arch geometry as the overhang is being scanned. The area on the overhang is significantly hotter compared to the standard scanning a few hundred layers later. The heat intensity locates areas of possible defects, such as rough surfaces and porous defects. This feature's emissivity inside the FLIR API is set to 1.0 for relative intensity. All the frames captured within the same layer are looped over to find the maximum value in each pixel for each layer. The maximum values for each layer are then merged into a single image and mapped to the 3D mesh file.

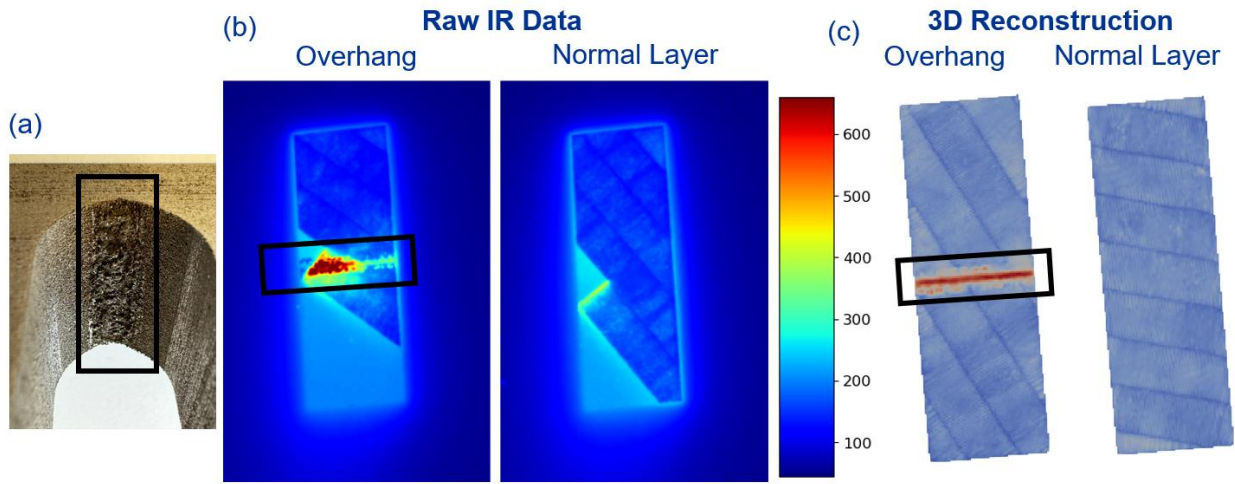

Figure 9: Comparison of heat intensity map on a layer with and without overhang nearby: (a) Image of the overhang of an arch showing a rough surface finish on the down-facing surface, (b) snapshot of the raw IR image of a layer of the (left) overhang and (right) bulk during printing, (c) corresponding heat intensity map on the (left) overhang and (right) bulk.

Next, the QTA block is used to demonstrate another application of the heat map for identifying process anomalies. Designed to evaluate the mechanical properties in various orientations, the two tensile bars tilted at a 45° angle encountered unexpected bending due to recoater blade interference, leading to a build failure. Since the EOS system was not capable of detecting the large displacement of the tensile bar, the laser kept scanning the regions where the tensile bar was supposed to be located in STL file, but since no solid was there to conduct the heat away quickly, the heat intensity increases substantially in the heat map as shown in Figure 10.

The process anomalies shown above often evade detection, allowing builds to progress unchecked until completion. Moreover, internal cavities and lattice structures pose additional challenges for post-build inspections. The heat map demonstrates the efficacy of the proposed IR data processing in providing valuable insights into process anomalies, particularly as additive manufacturing scales up to incorporate larger, more intricate shapes, where traditional inspections prove costly and time consuming.

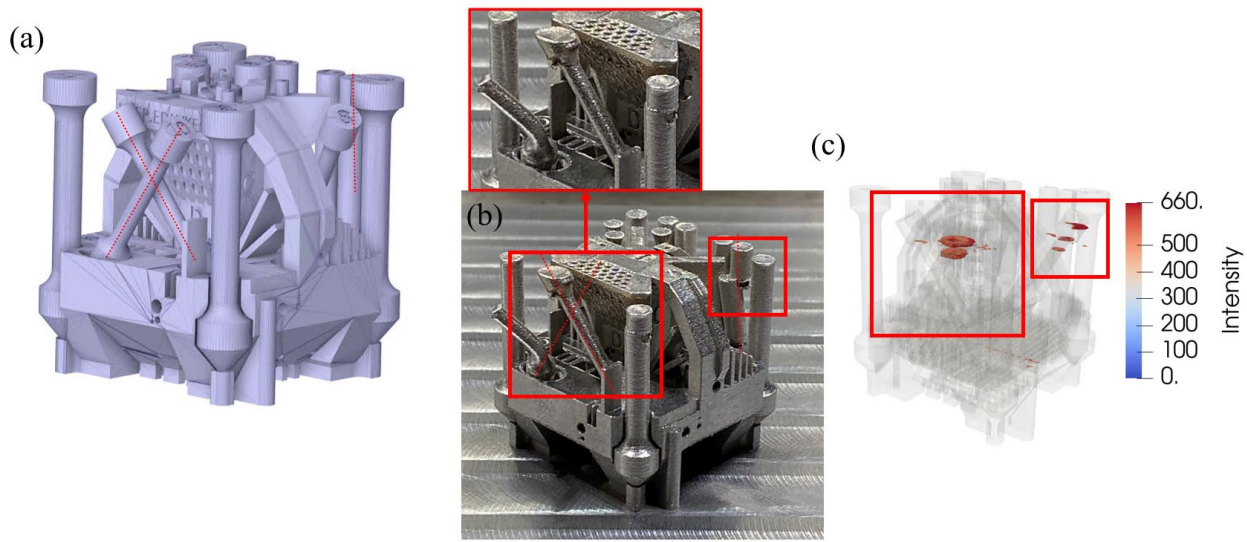

Figure 10: Demonstration of using IR intensity to identify process anomalies when printing a complex part: (a) A 3D view of the QTA block [23], (b) build failure where two angled tensile bars and one column (colored red) were knocked over by the recoater blade, (c) 3D reconstructed heat map of the QTA block highlighting intensity over 500 where the anomalies occurred.

### 4.3 Scan Order

The scan order algorithm tracks the laser scanning path in a layer, which is crucial for defect detection and process simulation as this information is usually not provided in commercial L-PBF systems. The algorithm provides information for the frame numbers for the spatter, melt pool area, cooling rates, pre-deposition temperature, and maximum preheat temperature. It is also an excellent separation point for changing the emissivity from 0.63 to 0.21 between the powder and as-printed surfaces for accurate temperature measurements as well.

Calculation of the scan order occurs concurrently with the heat map calculations. In each iteration, the algorithm loops through each frame, determining the maximum value between the existing heat map image and the new frame. For each pixel, when the new heat map exceeds the previous intensity, the frame number is updated in the laser location matrix using the NumPy package *Numpy.where()* such that the frame number at which the maximum heat intensity is recorded. Thus a 2D map showing this specific frame number for each pixel in a layer can be constructed to provide the scan order, where an example is shown in in Figure 11. It can be clearly observed that the laser scanning starts in the lower left side of the square and proceeds from one stripe to the next in a bi-directional manner toward the upper right side. Note that the stripe orientations and boundaries can be clearly discerned. Given the limitation of the 30 fps of the camera, recovering individual scan tracks poses a challenge as the laser would have scanned 32 mm in length between frames or ~3 scan tracks assuming a scan speed of 960 mm/s and a stripe width of 10 mm. However, the overall accuracy of the scan order and stripe configuration remains accurate.

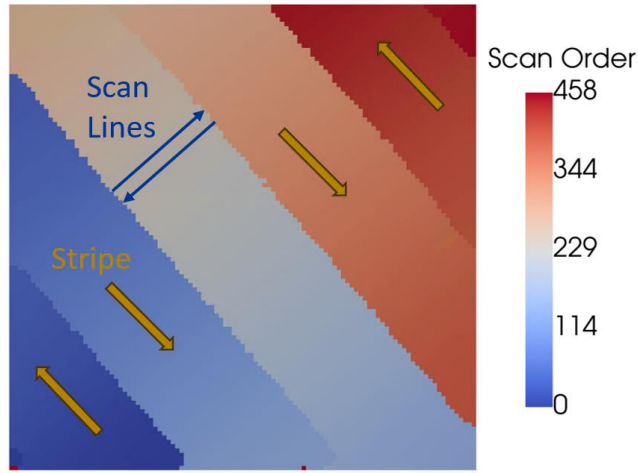

Figure 11: Demonstration of scan order algorithm processing IR data taken on a square block with stripes. The laser scanning starts in the lower left side of the square and proceeds from one stripe to the next in a bi-directional manner toward the upper right side. This algorithm clearly captures the stripe boundaries and orientations.

### 4.4 Local Pre-deposition Temperature

The local pre-deposition temperature is the local temperature right before the laser scans the location of interest and can affect the melt pool size and spatter generation rate. As parts are built throughout the layers, the temperatures can increase due to residual heating at an overhang or within a stripe, as well as in areas where the stripe boundary conditions interact with the next scanning location. In our implementation, the local deposition temperature at a given location is calculated 10 frames ahead of the maximum temperature within the melt pool as illustrated in Figure 12(a). 10 frames ahead of the scanning was found to be the optimal value providing data just before lasering without obtaining high values from the previous scan track. The feature shows the local pre-deposition temperature varying by over 220 °C throughout the layer.

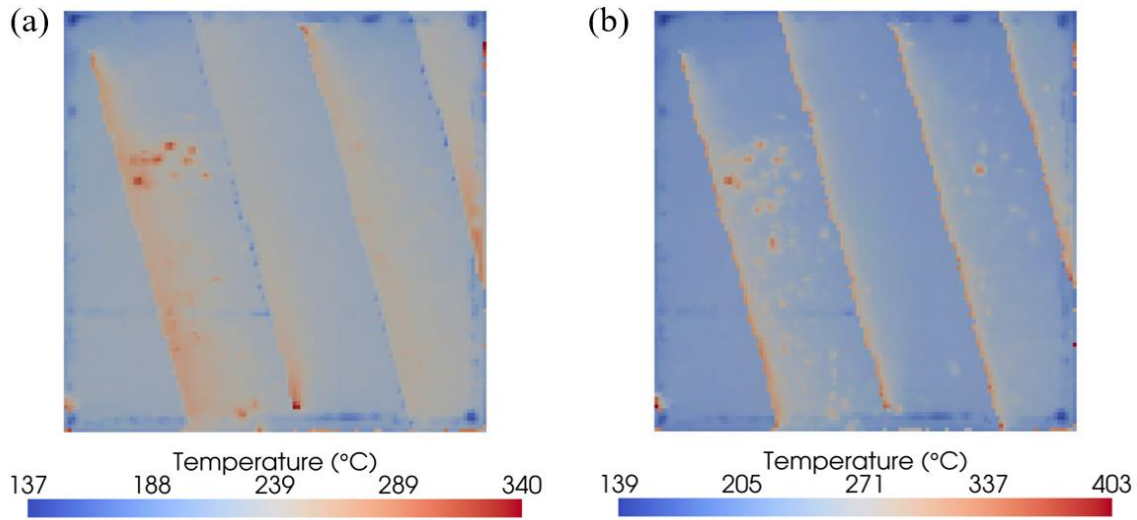

Figure 12: Comparison of the local pre-deposition temperature and maximum local pre-deposition temperature: (a) local pre-deposition temperature taken at 10 frames before the laser scanning and (b) Maximum pre-deposition temperature over all the past frames until 10 frames before laser scanning.

**4.5 Maximum Pre-deposition Temperature**

The maximum pre-deposition temperature is very similar to the local pre-deposition temperature; however, the algorithm looks for the maximum temperature for a pixel prior to the arrival of melt pools. Note that the global maximum temperature in a pixel is within the melt pool, and hence the maximum pre-deposition temperature is determined by searching through the data 10 frames before the global maximum temperature occurs in that pixel. In the algorithm, only the pixels with a frame number less than the scan order frame number minus 10 are selected using *numpy.where()*, and then a heat map *numpy.maximum()* is used to compile all the frames with the maximum temperature in all the pixels in the layer into a single composite image. An example of this feature is shown in Figure 12(b). This feature can be employed to detect hot spatters that land along future scanning path in the same layer. For example, isolated hot spots can be clearly observed in the first and third stripes from the left side of the square, and they are presumably caused by hot spatters landing on the powder surface. When spatter originates from another part, the local pre-deposition feature may overlook the spatter landing, as it has time to cool down. In contrast, the maximum pre-deposition temperature records all instances of spatters, even in scenarios involving multiple parts in the build.

## 4.6 Spatter Generation and Spatter Landing Locations

Spatter induced by the laser can land on the part that has yet to be scanned. This can be detrimental to the part as the hot spatter can be larger than the surrounding powder [4]. The other powder particles can also adhere to the spattered particle to form a larger aggregate, which may lead to a lack-of-fusion defect as the laser is not able to melt the entire aggregate. Detecting the spatter landing location poses challenges such as noise around the melt pool, variation in cooling rates, and part-to-powder interface, as depicted in Figure 13(a). In the proposed algorithm, a filter mask is first generated to encompass the melt pool and previously scanned areas, reducing data noise. Second, filtering identifies sharp temperature changes indicative of spatter or laser activity, see Figure 13(b). Upon combining these filters, only spatter points outside the mask are considered valid as shown in Figure 13(c).

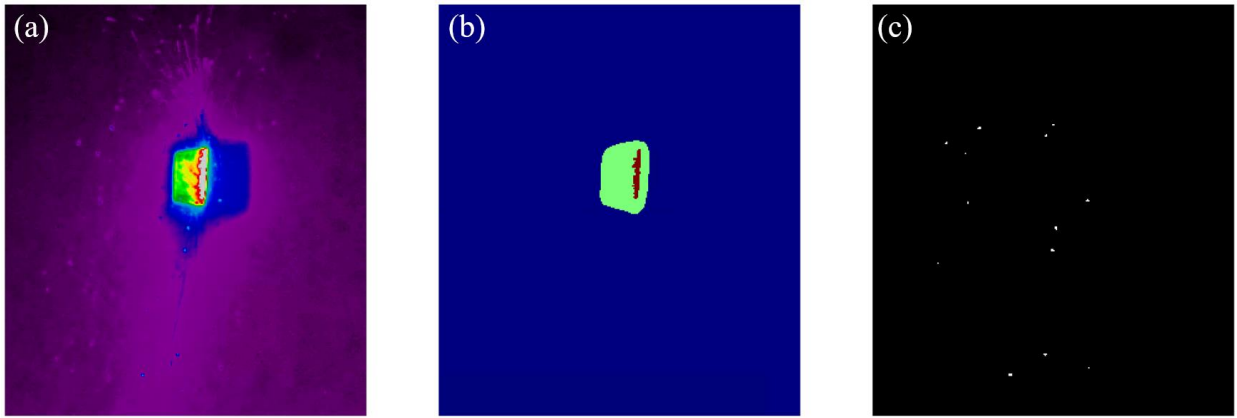

Figure 13: Spatter detection from the IR monitoring: (a) Raw IR data while scanning, (b) scanning filter algorithm with green representing areas previously scanned and red representing the current melt pool area, (c) further spattering filtering yields an image with easily identifiable spatters.

The detailed implementation of these filters is discussed here in details. The filter mask is first calculated as follows: The Gaussian gradient filter using the SciPy function *ndimage.gaussian_gradient_magnitude(sigma=3)* is utilized to find the areas with a high rate of temperature change where *sigma* is the standard deviations of the Gaussian filter. By locating the melt pool area, the laser location and previous scanned location are captured by a mask to only capture spatter landing before laser scanning. The function *threshold_multiotsu(classes = 2)* is then utilized to segment the image into 2 classes of images being the background and high temperature gradient areas, then digitized to 0 or 1 using *np.digitize()*. The result of this filter generates the mask around the laser and previously scanned locations as illustrated in Figure 13(b). The red region, laser location, is added as a reference to highlight the scanning region compared to the masking area.

Next, the spatter filtering technique is calculated as follows: Once the areas scanned and laser location are known, *ndimage.gaussian_laplace(sigma=1), Laplacian of the Gaussian (LoG),* emphasizes the local spatters, part boundaries, and the melt pool. Those areas are filtered out using the previous results, which isolates the spatter particles that have landed on the part yet to be scanned. The SciPy *scipy.ndimage.label()* function is utilized to find the spatters which are defined as clusters of pixels that are all connected to each other within the cluster. The algorithm finds each spatter cluster in the binary image in order to calculate the size and number of clusters (or spatters) in the image. Using spatter clusters ensures that spatter is not overcounted if a spatter is larger than a single pixel. As illustrated in Figure 13(c), the number of spatters is counted in each frame and the spatter count is assigned to the current laser location. The spatter landing location is also mapped to the area of the part they landed in for use in subsequent statistical analysis or machine learning for defect detection.

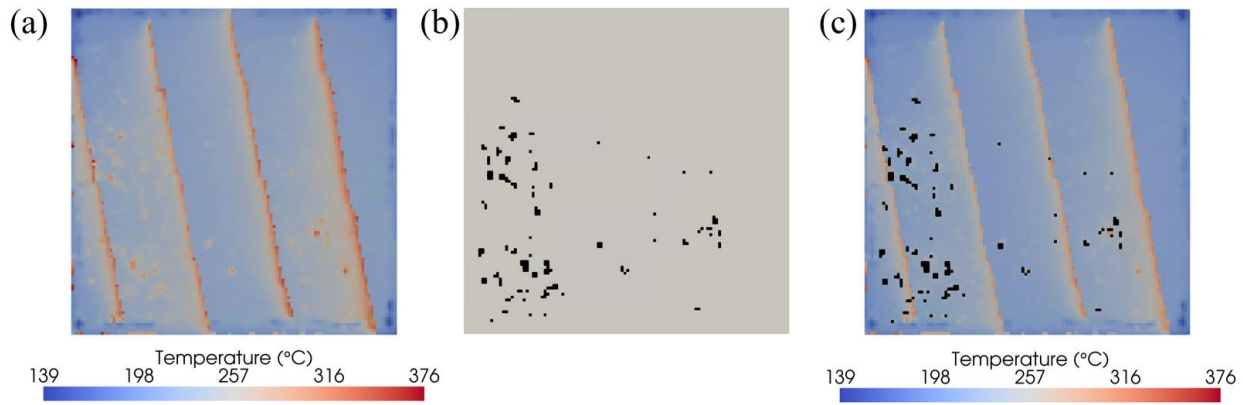

Figure 14: Comparison between the maximum pre-deposition temperature and the spatter landing detection algorithm: (a) the maximum pre-deposition temperature, (b) predicted spatter landing locations, and (c) predicted spatter landing locations overlaid onto the maximum pre-deposition temperature, showing good correlation.

It is difficult to verify the spatter landing locations using ex-situ measurement such as surface topography scanning since hot spatters often merge well with the part itself to increase its surface roughness. Thus we employ another IR feature, the maximum pre-deposition temperature, to cross-verify the spatter landing locations, see Figure 14. Note that except the stripe boundaries, the high-amplitude spots in the maximum pre-deposition temperature map also indicates spatter landing locations (Figure 14(a)). By overlapping this map with the spatter landing location map (Figure 14(b)), we can see that most of the high-amplitude spots in the former map coincide with the spatter landing locations. Throughout the build, hundreds of hot spatters are generated, but not all of them lead to defect formation. Hence this feature can be employed along with other features to improve the fidelity of defect detection using machine learning.

The proposed spatter landing feature has the potential to identify defects introduced by the landed spatters on the unscanned part. To demonstrate this possibility, a 19.44×19.44×5 mm block was printed in IN718 using the EOS system with in-situ IR monitoring during printing. The block was repeatedly grinded, polished, and imaged using an optical microscope at a distance of ~8 µm between successively imaged slices using a Robo-Met machine (UES, Inc, Dayton, OH). The Robo-Met machine uses an autofocus to calculate the removed material per slice which is then used to correspond to the layer number. Figure 15 shows a relatively large spatter identified by the proposed spattering landing algorithm on Slice No. 117 from the top surface or build layer 101, where the spatter landing location coincides with a 60×47µm pore observed in the optical microscopy in the same slice.

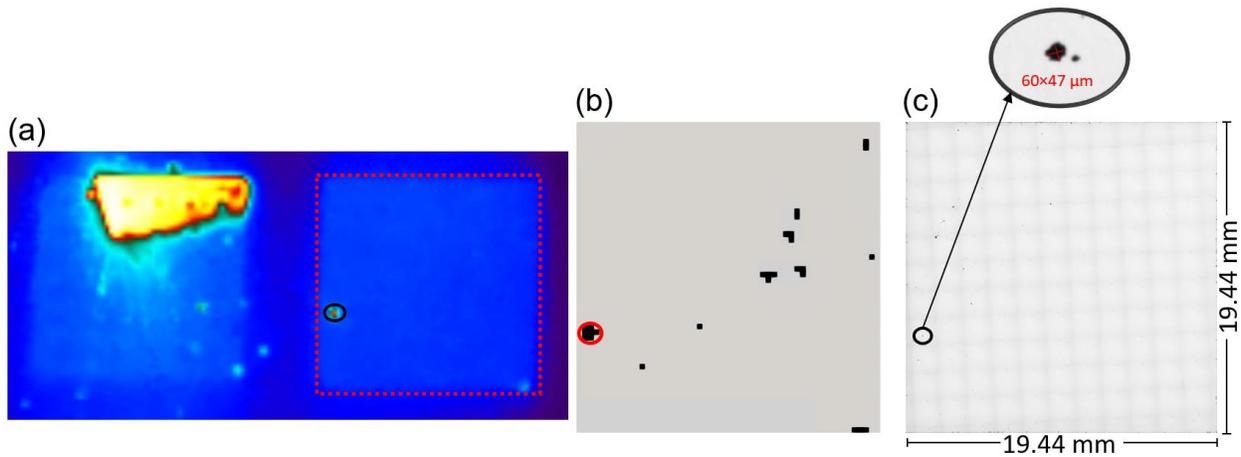

Figure 15: Evidence of the spatter landing feature causing a large pore during printing of a square block of size with a hole inside: (a) raw IR data illustrating a spatter (circled in black) landing onto the part while scanning a different part, (b) Results from spatter landing algorithm showing the same spatter circled in red, and (c) optical image showing a pore of size 60×47 µm near the spatter.

**4.7 Melt pool area**

Knowing the general trend of the laser melting area can be beneficial in monitoring the build quality and laser process parameters. If there are abnormalities caused by large spatters, overhangs, or short scan lengths, the melt pool size will change. Even though the IR monitoring employs a voxel size of 360 microns, 30 fps, and a maximum temperature of 660 °C, it can still capture the extent of the melt pool region by thresholding the temperature. Hence, this IR feature attempts to capture the relative melt pool area rather than the absolute area. The emissivity value of the material is typically set to provide accurate temperature values; however, the value can be changed to help highlight large changes in temperature, creating more efficient thresholds. In this case, a lower emissivity of 0.1 is set on the image to make thresholding more efficient by increasing the relative temperature inside the melt pool to the maximum allowed temperature value. The number of pixels above a threshold of intensity that represents temperature above 660 °C is then counted for every frame inside the layer. The total number of elements above the threshold is mapped to the laser location as illustrated in Figure 16. The *numpy.where()* function is utilized to locate the number of pixels above the threshold, whereas the length of the result provides the number of pixels. Consistent with scan-resolved thermal process simulation [29], the area where the scan first starts typically has a much smaller melt pool area compared to the rest of the part. Areas where the scan length varies can greatly affect the relative melt pool area.

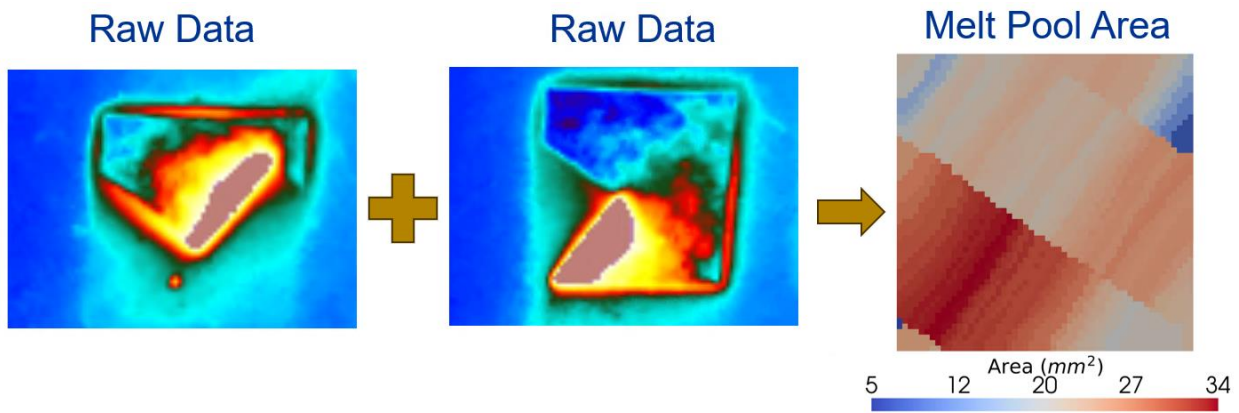

Figure 16: Raw IR data with a low emissivity to highlight the highest areas around the melt pool. The algorithm counts the number of pixels above this threshold and maps it to the laser location in the melt pool area feature.

**4.8 Relative Cooling Rate**

The relative cooling rate contains information regarding local thermal conductance, which can be used to detect processing defects. Since the frame rate and pixel resolution are not very fine, this feature is computed over a relatively long duration (one second). Specifically, the relative cooling rate is calculated using the difference between the maximum temperature and 30 frames (1 second) after the maximum temperature to calculate the rate of cooling. The cooling rate is calculated based on the laser location providing a consistent metric across all pixels. Figure 17 depicts the cooling rate results for the QTA block. In Figure 17(a), the cooling rate for Layer 840 is displayed for the standard layer before the large displacement of the tensile bar oriented at a 45 ºC, shown in Figure 10, accompanied by a heat map in the lower left corner. Figure 17(b) presents Layer 841 when the tensile bar has failed, exhibiting a notably lower cooling rate compared to the previous layer. This observation aligns with the corresponding heat map feature. Layer 840 exhibits a standard temperature range on the heat map, in contrast with Layer 841 where a significant increase is evident.

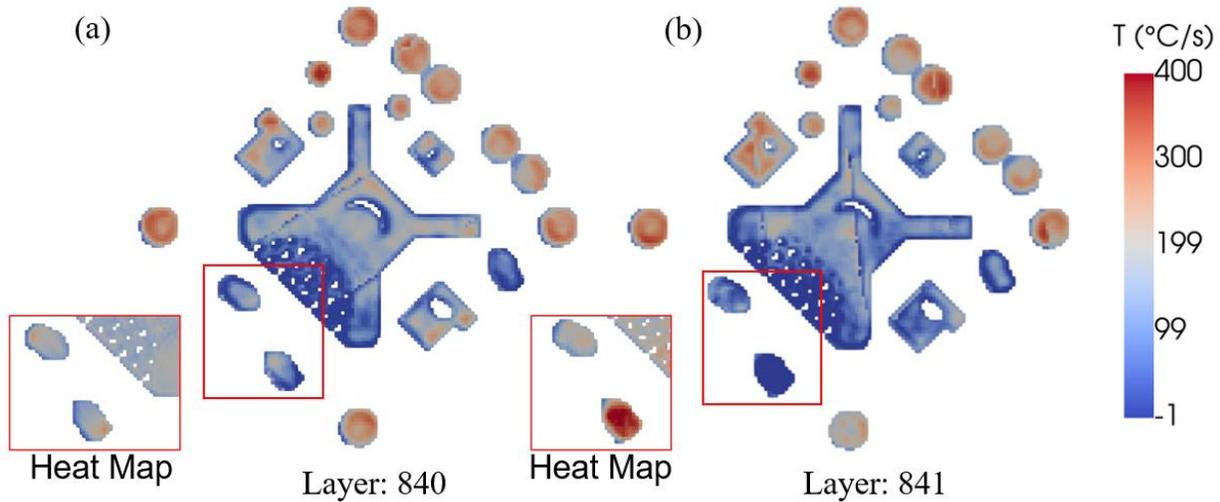

Figure 17: Cooling rates in two different layers of the QTA block before and after the tensile bar is displaced in the build: (a) On Layer 840 before the tensile bar is displaced. (b) On Layer 841 after the tensile bar is displaced. The areas in red highlight the large change in cooling rates (and heat intensity) before and after the large tensile bar displacement.

### 4.9 Laplacian of temperature

Throughout the build, there are many instances where the recoater blade will not fully cover the entire part. Sometimes, there can be streaks on every layer due to an issue with the recoater blade. Other times, the part may deform upwards enough to stop a fresh layer of powder from being laid. Wang et al. [17] pointed out the temperature drop due to damage from the recoater blade recoating [21]. The temperature drop is caused by the lower emissivity value of the as-printed surface compared to the powder surface. Using the interpass temperature only, it is difficult to detect any recoating issues because temperatures can greatly vary throughout the build due to scanning strategy and local geometry, as shown in Figure 8. However, if the Laplacian (or the second spatial derivative) of the temperature field is used, these recoat defects can be more clearly identified as shown in Figure 18(a). The high Laplacian values in the figure correspond to the locations with high pixel color contrast in an image (see Figure 18(b)) captured by a high-resolution camera (Basler, Ahrensburg, Germany) with a resolution of 132 microns per pixel. The brighter areas in the Basler image correspond to areas without powders, providing evidence of incomplete powder recoating. The Laplacian helps to find pixels with a high rate of temperature change compared to surrounding pixels. As presented in Section 2, the emissivity values of as-printed surface and powder are respectively 0.63 and 0.21, which cause a significant temperature difference in areas without new powder. Even though the temperatures on this build reached over 500 °C, the powder spread value is approximately zero inside the part unless there is a recoat issue, in which case the Laplacian value will increase because of the large rate of change in temperature. The Laplacian is computed by the SciPy function *image.gaussian_laplace()*, followed by an image correction using OpenCV before mapping to the geometry.

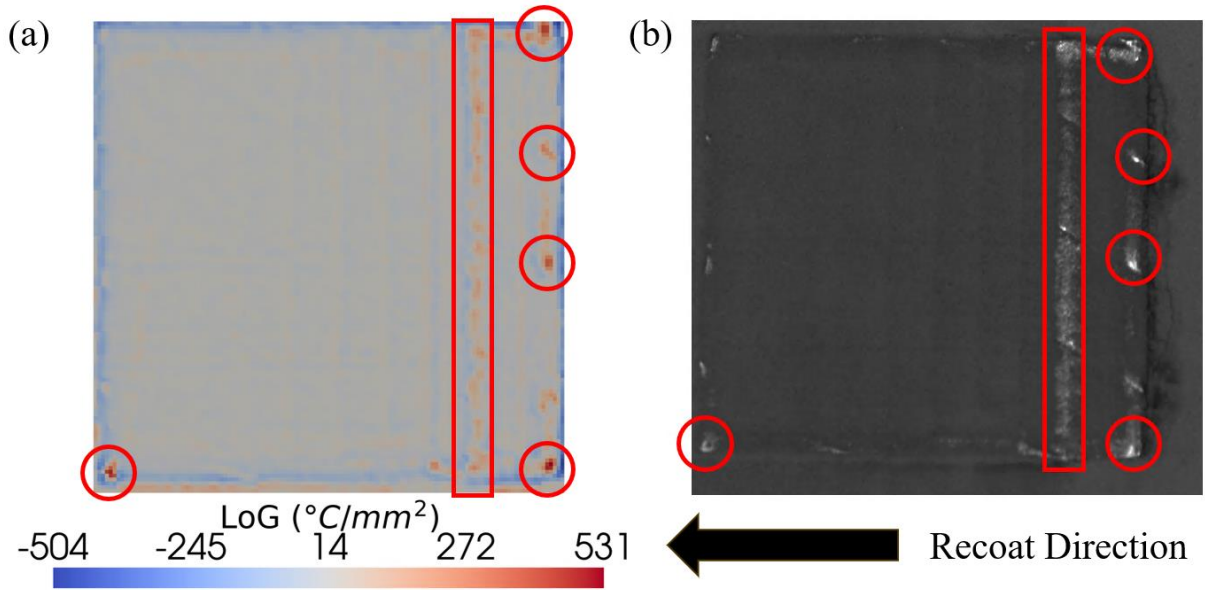

Figure 18: Detection of powder spread anomalies using Laplacian of the temperature feature and validation using a high-resolution camera with a resolution of 132 microns per pixel: (a) Laplacian of the Gaussian (LoG) temperature field with high values circled or boxed in red and (b) image captured by the high-resolution camera on the same layer with high color contrast regions circled or boxed in red indicating incomplete powder recoating.

**4.10 As-printed Laplacian**

As shown in Section 2, the as-printed surfaces exhibit large variations in emissivity primarily due to the surface roughness variation over a surface. As the temperature surrounding each pixel should be relatively constant after a few seconds of cooling, the temperatures should not change much between neighboring pixels. There are some sharp changes that are typically caused by the surface roughness of the part. Figure 19 illustrates a typical case where there is a higher gradient near the stripe boundaries caused by the increased roughness in those areas. The Laplacian feature provides consistent values throughout layers, whereas the temperature values can change significantly from layer to layer. *image.gaussian_laplace(sigma=1)* provides much more uniform values even when temperatures change throughout the layers.

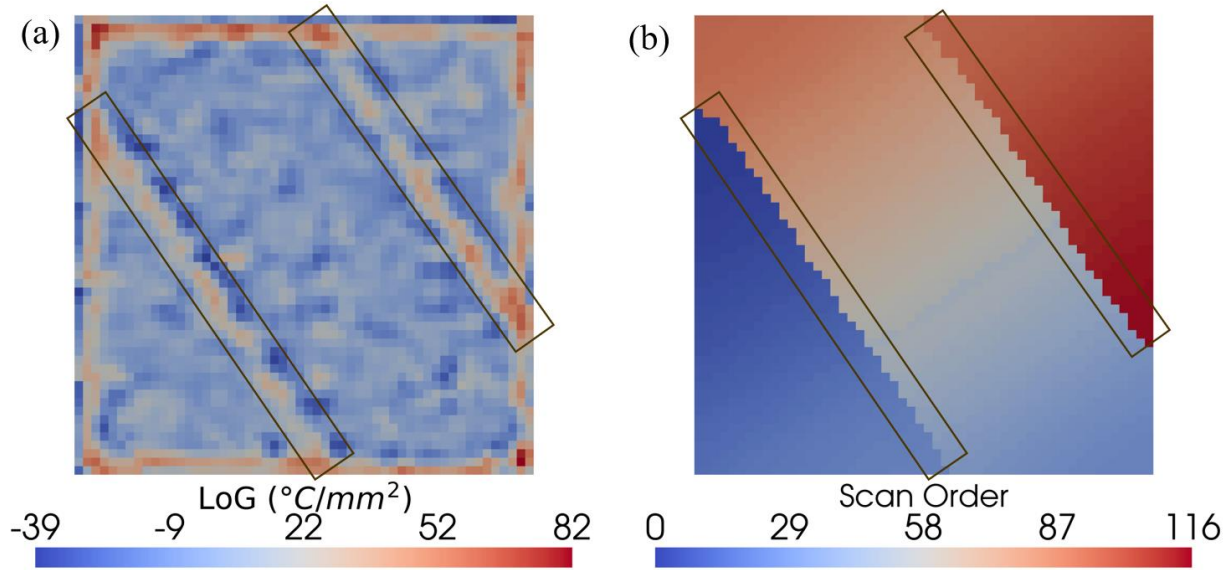

Figure 19: Illustration of the temperature Laplacian as compared with the scan order employed to print a block: (a) Laplacian of the Gaussian (LoG) temperature and (b) scan order. A correlation is found at the stripe overlaps where the Laplacian shows higher values than other areas.

## 5 Conclusion

This work proposes a new approach to accurately map in-situ IR monitoring data collected during LPBF processing to a 3D geometry with angle correction. To show the robustness of the algorithm, the QTA block is employed to show that the IR data is mapped onto the complex geometry correctly and the storage requirement is reduced by 99% in the process. Also, the algorithms for processing different IR features are introduced and several of them are demonstrated to correlate with process defects and anomalies. The following conclusions can be drawn:

- Through calibration using a thermocouple, the interpass temperature obtained by the IR camera has a 5.0 °C maximum average error and a standard deviation of 3.3 °C on powder surface, and a 20 °C maximum average error and a standard deviation of 25.2 °C on as-printed surface.
- The heat intensity and cooling rate features are both capable of detecting large displacement of parts in a component and change in surface roughness.

- The maximum pre-deposition temperature and spatter landing location features can be associated with pores observed in the optical imaging experiment.
- Large values of the Laplacian of temperature can be correlated with incomplete powder recoating, which is often caused by recoater blade interference with the part.

Future work will utilize the interpass temperature feature to calibrate and validate thermal process simulation models, thereby enhancing the accuracy of temperature predictions. Additionally, the extracted IR features will be utilized in machine learning algorithms to detect porous defects and microstructures as part of qualification. There is also potential to extend the proposed techniques to enable real-time defect detection and repair during processing. The proposed in-situ mapping technique and IR features pave the way for improved accuracy in thermal-mechanical simulations, quality control, processing monitoring, and optimization within the AM process.

**CRediT authorship contribution statement**

**Shawn Hinnebusch:** Methodology, Conceptualization, Investigation, Data curation, Software, Writing, Review, and Editing. **David Anderson**: Data Curation and Investigation. **Berkay Bostan**: Investigation. **Albert To:** Conceptualization, Funding Acquisition, Supervision, Project Administration, Writing and Editing.

**Declaration of Competing Interest**

The authors declare that they have no known competing financial interests or personal relationships that could have appeared to influence the work reported in this paper.

**Declaration of Generative AI and AI-assisted technologies in the writing process**

During the preparation of this work the author(s) used ChatGPT by OpenAI in order to proof read paragraphs for grammatical errors and clarity in early drafts. After using this tool/service, the author(s) reviewed and edited the content as needed and take(s) full responsibility for the content of the publication.

**Acknowledgements**

This project was sponsored by the Department of the Navy, Office of Naval Research under ONR award number N00174-23-1-0011. Any opinions, findings, and conclusions or recommendations expressed in this material are those of the author(s) and do not necessarily reflect the views of the Office of Naval Research.